\documentclass[11pt,a4paper]{scrartcl}

\usepackage[english]{babel}
\usepackage[letterpaper,top=2cm,bottom=2cm,left=3cm,right=3cm,marginparwidth=1.75cm]{geometry}
\usepackage[ backend=biber, style=alphabetic, url=true ]{biblatex} 
\usepackage{todonotes}
\usepackage[T1]{fontenc} 
\usepackage{lmodern}
\usepackage{amsmath, amssymb}
\usepackage{graphicx}
\usepackage[colorlinks=true, allcolors=blue]{hyperref}
\usepackage{xurl}
\usepackage{enumitem}
\usepackage[utf8]{inputenc}
\usepackage{tikz}
\usepackage{pgf-pie}

\title{LLM-Based Policy Compliance Verification of Loan Applications}
\author{Florentina Voboril\\ 
Technische Universit\"at Wien\\ 
Erste Bank\\
{\small \texttt{fvoboril@ac.tuwien.ac.at}}
}
\date{}

\begin{document}
\maketitle

\begin{center} 
\emph{This work was carried out during an internship at Erste Bank der oesterreichischen Sparkassen AG (Erste Bank) in cooperation with TU Wien.} 
\end{center} 

\vspace{1cm}

\begin{abstract}
Banks rely on extensive internal policy frameworks to ensure that corporate lending decisions comply with regulatory requirements and internal risk standards. Reviewing loan applications against these policies is a largely manual and knowledge-intensive process, which can lead to overlooked policy violations or exceptions. This report presents a prototype of an LLM-based policy checker that automatically verifies loan applications against internal lending policies.

The proposed approach follows a two-step architecture. First, compliance-relevant rules are automatically extracted from policy documents and converted into structured, machine-readable representations. Second, loan applications are evaluated against the extracted rules to identify policy exceptions, non-compliant findings, missing information, and cases requiring manual review. 

Evaluation results demonstrate that the rule extraction process is highly stable across multiple runs and consistently captures rules identified as critical by a domain expert. When evaluated on 11 real loan applications with known policy exceptions, the system successfully detected the majority of these exceptions and identified additional compliance issues that had previously been overlooked during manual reviews. While the approach produces false positives and cannot fully replace human judgement, the results indicate that LLM-based policy compliance verification can provide valuable support in the loan application verification process by highlighting potential compliance risks.
\end{abstract}

\section{Introduction}

Banks use many internal policies and guidelines when deciding whether to approve a loan. These policies define what customer information is required, which conditions must be met, and which policy exceptions must be flagged. A policy exception is a formally defined exception from a policy requirement that may be permitted under specific conditions but must be explicitly documented. Policy documents are defined across different institutional levels and cover different types of customers and financing products. Further, they are updated regularly. As a result, it can be difficult to ensure that every loan application complies with all relevant policies.

This report focuses on corporate lending, that is, lending to companies. Here, a loan application refers to the documentation prepared within the bank to support a lending decision. These documents are primarily prepared by account managers who must know which policies apply, understand their requirements, and correctly include them in the application. Because this largely manual process depends heavily on individual knowledge and expertise, requirements can sometimes be overlooked.

Current advances in Large Language Models (LLMs) provide new opportunities to support automated compliance checks. Their ability to interpret regulatory requirements, identify relevant information across documents, and reason over unstructured texts makes them a promising tool for automated policy analysis and compliance verification.

Recent research has explored the application of LLMs to financial regulation and compliance tasks. Prior work has shown that LLMs can interpret complex regulatory requirements and translate them into structured representations suitable for automated processing and verification \cite{cao2024financial}. Other studies have investigated the extraction of obligations and policy requirements from regulatory documents to support compliance reasoning and question answering \cite{baldwin2026kg}. In parallel, LLMs have increasingly been applied to financial decision-making and loan assessment, demonstrating considerable potential while also highlighting the importance of human oversight and governance \cite{bowen2025mortgage}. Building on these developments, this work investigates whether LLMs can support the review of corporate loan applications by automatically verifying compliance with internal lending policies.

We present a prototype for an LLM-based policy checker for corporate loan applications at Erste Bank. The proposed system automatically extracts rules from internal lending policies and evaluates loan application documents against these rules to identify policy exceptions, missing information, and potential violations.

The remainder of the report is organized as follows: Section~\ref{policies} provides background information on the relevant policy framework. Section~\ref{implementation} describes the implemented system, including the preprocessing of relevant documents, the system architecture, and the agent design. Section~\ref{evaluation} evaluates our system on real loan applications, while Section~\ref{conclusion} summarizes the main findings, outlines the limitations of the proposed approach, and discusses directions for future work.

\section{Loan Applications and Policies}
\label{policies}

Loan applications are comprehensive documents that contain various types of information, including customer information, financial statements, purpose of the application, collateral details, risk assessments, and comments from the account manager. The corresponding lending rules are distributed across multiple policy documents and organizational levels. While this internship focused on corporate lending at the Erste Bank Österreich (EBOe) level, the EBOe Corporate Lending Principles are mainly a cover sheet that directly adapts the Group Corporate Lending Principles, with only a few locally adjusted paragraphs. The underlying Group policy contains 90 numbered paragraphs and seven annexes, of which the Country Specifics and the Policy Exceptions annexes were particularly relevant for this project. These annexes introduce additional complexity since country-specific values have to be looked up for the corresponding thresholds and policy exceptions must be identified and documented. Furthermore, the policy references 38 other related documents and references covering topics such as real estate financing, renewable energy financing, collateral management, rating regulations, and approval procedures, which were considered out of scope. Another challenge is that not all policy statements are directly related to information that must be contained in a loan application. Some principles regulate internal processes, governance requirements, or approval workflows. Others regulate monitoring activities, documentation standards, or responsibilities of account managers and risk managers. As a result, an important task of the internship was to determine which policy rules could be evaluated using information available in the loan application and which relied on data stored in external systems or documented in separate processes. This distinction is essential for the design of the policy checker.

\section{Our implementation}
\label{implementation}

\subsection{Platform \& Data Protection}

The policy checker is built and executed on Databricks, a cloud-based platform for data processing, analytics, and AI workloads. The agent framework used is \texttt{pydantic-ai} (v2.38.0), a Python library for building LLM-powered agents with structured, type-safe inputs and outputs based on Pydantic models.

We use Claude Opus 4.8 (\texttt{databricks-\allowbreak claude-opus-\allowbreak 4-8}) as the LLM, an Anthropic model hosted within the Databricks serving infrastructure. Since the pipeline processes financial and commercially sensitive information from real clients, data protection is of particular importance to this project. Databricks contractually guarantees that ``\textit{Model Serving does not use user inputs submitted to the service or outputs from the service to train any models or improve any Databricks services}''~\cite{databricks_model_serving}. As a result, the contents of loan applications and the extracted rules remain confidential and are not subject to human review by the provider.
Further, all requests are authenticated and logically separated from those of other customers, and data is encrypted both during transmission and while stored within Databricks' infrastructure~\cite{databricks_model_serving}.
In addition, all processing takes place within the European Union. The Databricks workspace is hosted in the Azure West Europe region, and Foundation Model API requests remain within Databricks' European infrastructure~\cite{databricks_fm_compliance}.
These technical and contractual safeguards allow the processing of real, non-anonymized client data while remaining compliant with GDPR requirements and Erste Group's internal data governance policies.

\subsection{Input Documents and Preprocessing}

The policy checker processes information from multiple files with different document types: The Corporate Lending Principles (DOCX), the loan applications (PDF), and two annexes (PDF and Excel). Each of these requires its own parsing strategy. Our parsing pipeline is built on a custom \texttt{document\_processing} library that wraps Azure Document Intelligence for PDFs and handles DOCX natively. The goal of the parsing step is to convert unstructured, heterogeneous source documents into clean, formatted strings that the LLM agents can process.

\subsubsection*{Corporate Lending Principles (DOCX)}

With 50 pages, the Group Corporate Lending Principles guideline is the primary input document of this project. It consists of 90 numbered paragraphs, which we want to split up into individually accessible, machine-readable paragraphs. First, the \texttt{document\_\allowbreak processing} library parses the DOCX into an HTML representation. This representation allows us to identify and elevate the ordered list items that encode the individual paragraphs. Further, all content before the first paragraph, including the table of contents, is removed.

The preprocessing yields two outcomes. First, it produces a structural overview of the document, including all headings and the beginnings of each paragraph. This allows the agent to navigate through the document and look up definitions or cross-references on demand. Second, it returns a list containing each paragraph as an individual textual item. The three paragraphs where the EBOe Cover Sheet states differences from the Group policy were manually overridden.

\subsubsection*{Annex: Country Specifics (PDF)}

The Country Specifics annex contains a table with country-specific thresholds, such as the threshold at which covenant protection is mandatory. Covenants are contractual obligations imposed on a borrower, such as maintaining specified financial ratios. Further, the country specifics contain the permitted probability of default for each customer group, and the minimum collateral coverage by customer rating.  The PDF is parsed via the \texttt{document\_\allowbreak processing} library into an HTML table, which is converted into structured records in a subsequent step.

\subsubsection*{Annex: Policy Exceptions (Excel)}

The Policy Exceptions annex lists 16 predefined deviations from the principles that must be explicitly flagged. Each entry contains an exception identifier (e.g. \textit{``Policy Exception 1''}), a policy name (e.g. \textit{``Covenant protection''}) and the description of the exception. The Excel file is read with \texttt{pandas} and provides two outcomes: an overview listing all exceptions with their numbers and names, and a function to read a specific exception's description.

\subsubsection*{Loan Applications (PDF)}

A loan application is parsed from PDF into structured HTML parts by the \texttt{document\_\allowbreak processing} library. In a second step, each HTML part is converted to Markdown using the \texttt{markdownify} library. This preserves structural elements, like bold, italic, underline, headings, and table formatting, while stripping all remaining HTML tags. The result is a single plain-text Markdown string.

\subsection{Architecture \& Agent Design}

\subsubsection*{Two-Step Architecture}

\begin{figure}
    \centering
    \includegraphics[width=1\linewidth]{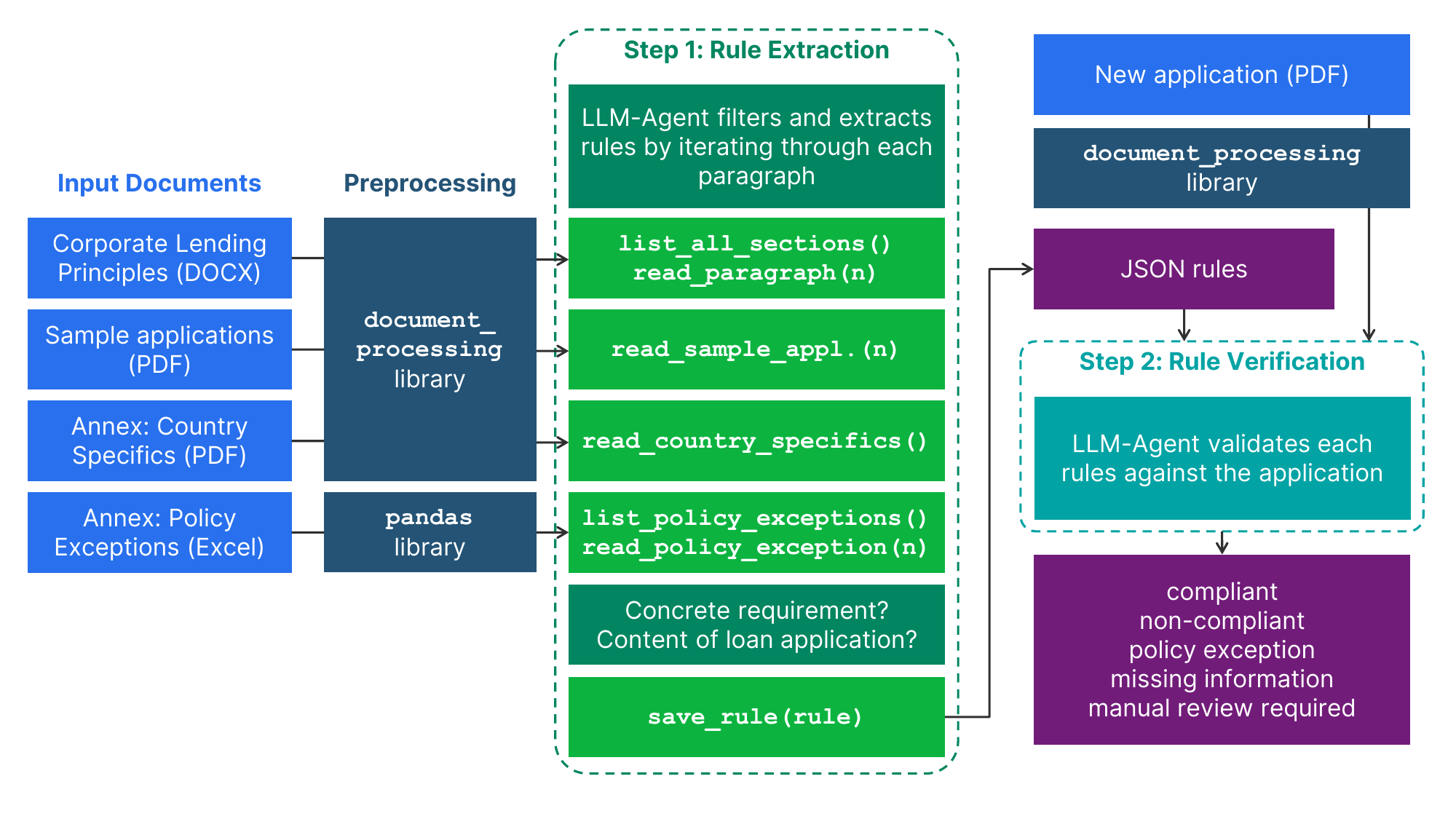}
    \caption{Two-Step architecture of the policy checker}
    \label{fig:diagram}
\end{figure}

The policy checker consists of two distinct stages. In a first step, concrete rules are extracted from the guidelines, and in the second step, loan applications are verified against these rules. This separation has multiple advantages: First, the rules do not have to be re-extracted for each loan application. The same set of rules can be reused until the guidelines are updated to a new version. Further, human experts can inspect, correct, or supplement the extracted rule set before it is used for automated verification. Also, when errors occur, it is easier to debug since one can determine whether the mistake happened during the extraction or the verification. An overview of the whole architecture can be found in Figure~\ref{fig:diagram}.

\subsubsection*{Step 1: Rule Extraction}

From the document preprocessing of the Corporate Lending Principles, we retrieve a list of all 90 paragraphs. We iterate through each paragraph and call the agent, passing the current paragraph and a detailed system prompt as input.

\textbf{Filtering logic.} Not every paragraph is expected to yield an extractable rule. To determine whether a paragraph should be processed, the system prompt instructs the agent to apply a two-step filtering procedure. First, the agent evaluates the following question: ``\textit{Does the paragraph contain a concrete REQUIREMENT that is directed at the CONTENT of a loan application?}'' If the answer is no, the paragraph is discarded. This step filters out paragraphs that merely define terms, describe procedures, or specify governance requirements for individuals or organisational units, such as approval authorities, four-eyes principles, or review obligations.
Second, the agent verifies the requirement against at least one of the provided sample loan applications. The purpose of this step is to determine whether the required information is structurally represented in a loan application. If no corresponding information can be identified, the paragraph is discarded. This verification step was introduced because the outcome of the first filter was not always clear. In some cases, the agent was unable to reliably determine from the principles alone whether a requirement was relevant to the content of a loan application. The additional verification against sample applications therefore serves as a practical relevance check.

\textbf{Output Structure.} Each extracted rule is stored as a JSON file with the following fields:

\begin{itemize}[noitemsep]
    \item \texttt{summary}: short summary of the rule in one or two sentences
    \item \texttt{description}: detailed description of the requirement
    \item \texttt{context}: explains detailed background on the rule, like technical terms and concepts 
    \item \texttt{example}: illustration of how this rule is fulfilled in the sample loan application
    \item \texttt{condition} (optional): precondition or prerequisite under which the rule applies
    \item \texttt{reference} (optional): reference to another framework or document
    \item \texttt{exception} (optional): circumstances specified in the rule under which its requirement does not apply
    \item \texttt{policy\_exceptions} (optional): relevant formally defined policy exceptions that
 account managers must explicitly document
    \item \texttt{paragraph}: number of the paragraph
    \item \texttt{source}: reference to the corporate lending principles
\end{itemize}

\textbf{Tools.} The agent has access to the following tools that allow it to navigate through the guideline, read sample loan applications, access the annexes, and finally save the extracted rule.

\begin{itemize}
    \item \textbf{\texttt{list\_all\_sections()}}: provides an overview of all sections of the corporate lending principles and their paragraph numbers.
    \item \textbf{\texttt{read\_paragraph(n)}}: retrieves the full text of the $n$-th paragraph. This is used by the agent to look up definitions or cross-referenced sections.
    \item \textbf{\texttt{read\_sample\_application(n)}}: returns the $n$-th sample application as full text. The list of available loan applications is included in the system prompt so the agent can select the most promising one without an extra tool call.
    \item \textbf{\texttt{read\_country\_specifics()}}: returns EBOe-specific thresholds from the corresponding annex. The system prompt instructs the agent to use this tool whenever a paragraph references country-specific values.
    \item \textbf{\texttt{list\_policy\_exceptions()/read\_policy\_exception(n)}}: list/read the 16 policy exceptions defined in the corresponding annex. The agent should consult these tools whenever the paragraph text references policy exceptions.
    \item \textbf{\texttt{save\_rule(rule)}}: saves the rule as JSON file.
\end{itemize}

\subsubsection*{Step 2: Rule Verification}

To validate a loan application, each extracted rule is assessed individually. For each rule, the agent receives the rule itself, the loan application, and a dedicated system prompt as input. The agent's task is to determine whether the loan application complies with the rule and to provide a justification based on evidence from the application. The output of this step is a JSON object containing the verdict and its justification. The possible verdicts are:

\begin{itemize}[noitemsep]
    \item \texttt{COMPLIANT}: the rule is satisfied by the loan application, including cases where the rule is not applicable because its precondition is not met (e.g., the total assets are below the relevant threshold)
    \item \texttt{NON-COMPLIANT}: the rule is applicable but not satisfied by the loan application.
    \item \texttt{POLICY EXCEPTION <name>}: the rule corresponds to a policy exception and the exception condition is met
    \item \texttt{MISSING INFORMATION}: compliance cannot be determined because required information is unavailable (e.g., a value needed for a threshold comparison is missing)
    \item \texttt{MANUAL REVIEW REQUIRED}: the rule cannot be assessed automatically and requires human judgement or manual verification
\end{itemize}

Finally, the individual rule outputs are aggregated and stored in a single text file.

\section{Evaluation}
\label{evaluation}

\subsection{Rule Extraction}

In order to check the consistency of the rule extraction step, we ran it five times. The extraction proved to be highly stable: out of 90 policy paragraphs, each run extracted 34–36 rules. 33 of the 37 distinct rules (89\%) were extracted in all 5 runs. The remaining 4 appeared in 1-4 runs. Notably, all 16 rules flagged as critical by an expert were extracted in all 5 runs, as were all rules directly referencing any of the policy exceptions -- in the latter case, the assignment to the correct exception number was also consistent across runs. Beyond these expected associations, the five runs identified 28 additional rule–exception links. Many of these turned out to be valid: some arose from indirect references, where one paragraph points to another that ultimately contains a policy exception, while others reflected exceptions that were contextually relevant to the rule even though the paragraph itself did not explicitly reference the annex listing the policy exceptions.

From the 5 runs, we selected the 36 rules extracted in at least 3 runs for our final rule set, which is evaluated in the next step. For each rule, the source run was chosen at random.

\subsection{Rule Verification}

We ran the rule verification agent on 11 loan applications with known policy exceptions, which were distinct from the sample applications used during extraction. In total, this resulted in $36$ rules $\times 11$ applications $=396$ rule checks. Out of these, 290 (73.2\%) were marked as compliant, 41 as policy exception, 38 as non-compliant, 25 as missing information, and 2 as manual review required. Figure~\ref{fig:verdicts} shows this distribution. However, it is important to note that some findings were marked within multiple rules in a single loan application, since related rules examine the same topic from different perspectives.
\newline

\definecolor{mygreen}{HTML}{0CB43F} 
\definecolor{myorange}{HTML}{FF6130} 
\definecolor{mypink}{HTML}{EB4C79} 
\definecolor{myblue}{HTML}{2870ED} 
\definecolor{myturquoise}{HTML}{02A3A4}

\begin{figure}[htbp]
    \centering
\begin{tikzpicture}
\makeatletter
\let\originalPieNumber\pgfpie@numbertext
\renewcommand{\pgfpie@numbertext}[1]{
  \ifdim#1pt=0.5pt\relax
  \else
    \originalPieNumber{#1}
  \fi
}
\makeatother
  \pie[
    text=legend,
    radius=3,
    color={mygreen,myorange,mypink,myblue,myturquoise},
    sum=auto,
    style=very thin,
    after number={\%}
  ]{
    73.2/Compliant (290),
    10.4/Policy exception (41),
    9.6/Non-compliant (38),
    6.3/Missing information (25),
    0.5/Manual review required (2)
  }
    \node[inner sep=1pt]
      (smallpercent) at (2.4, 0.3) {0.5\%};
    \draw[very thin]
      (smallpercent.south) -- (-0.9:2.4);
\end{tikzpicture}
    \caption{Distribution of the classification outputs across 396 rule checks.}
    \label{fig:verdicts}
\end{figure}
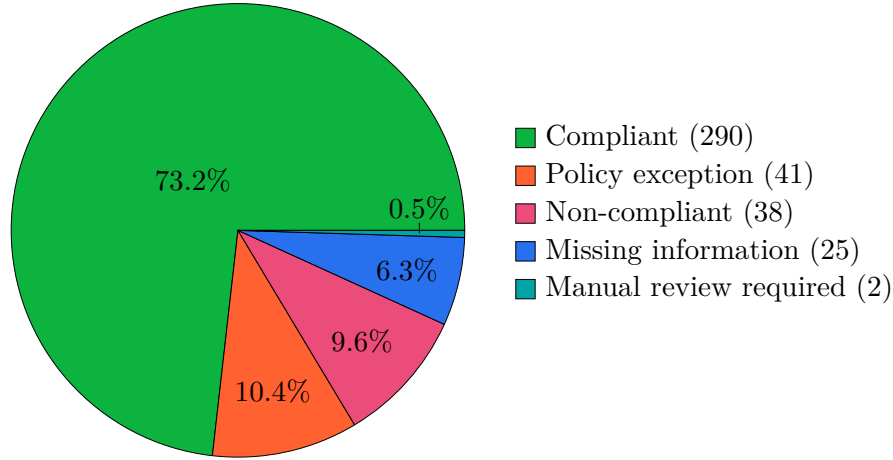

Out of the 13 previously known \textbf{policy exceptions} distributed over the 11 loan applications, the policy checker correctly marked 11 of them. For one of the missing policy exceptions, the LLM correctly recognized that the policy exception applies, but marked the rule as compliant since the existence of the policy exception was already pointed out in the loan application. For the other one, the LLM did not flag that covenants were missing. This was because at the end of the loan application, covenants from a previous loan were listed, and the agent could not determine whether they applied to the new or to the old loan.

On the other hand, the LLM correctly found policy exceptions that were not previously marked, such as missing collateral.

Since a complete classification into true-positive and false-positive findings would require a lot of manual effort and expertise, we decided to only examine selected findings together with an expert in greater detail and present the key highlights.

The LLM indeed found \textbf{non-compliant} rules in some of the loan applications:

\begin{itemize}
    \item For one loan application, the policy checker correctly determined that the required industry risk analysis was absent. This finding was accurate since the input data provided to the LLM did not include the full set of documents needed to satisfy the rule.
    \item One rule requires a full ESG (Environmental, Social and Governance) Assessment Questionnaire for SMEs above a certain turnover threshold, and an ESG Factor Heatmap for SMEs below it. The policy checker correctly identified the company's turnover and correctly concluded that no full questionnaire was required. However, the  ESG Risk Assessment Section was not filled out, and the missing ESG Factor Heatmap was correctly marked as non-compliant.
    \item The policy checker correctly identified that the required financial analysis was only partially available and contained significant gaps.
\end{itemize}

The rule checks marked with \textbf{missing information} mainly come from references to other documents, to which the policy checker has no access.

The two rules that are marked with \textbf{manual review required} stem from one rule that requires equity to be injected up-front and another one with unclear capital and cash flow coverage requirements.

The expert-reviewed examples suggest that the policy checker can highlight unmet requirements and policy exceptions overlooked during manual review. This suggests that the system may support review. However, the LLM is not always correct. Some rules were incorrectly flagged as non-compliant or as policy exceptions. The system exhibits a conservative bias and tends to over-flag rather than under-flag. While this produces false positives that must be reviewed manually, we consider this behaviour preferable to overlooking violations.

\section{Conclusion and Outlook}
\label{conclusion}

This internship explored the feasibility of using LLMs to automatically verify loan applications against internal lending policies. We developed a prototype policy checker that extracts machine-readable rules from policy documents and evaluates individual loan applications against these rules.

The evaluation demonstrated that the proposed approach is capable of identifying policy exceptions, missing information, and non-compliant findings that had previously been overlooked during manual reviews. In particular, the rule extraction step proved highly stable across multiple runs, and all policy-exception-related rules and rules identified as critical by a domain expert were extracted consistently. Furthermore, the verification agent successfully detected most known policy exceptions and produced findings that were confirmed as valid by expert review. These results indicate that LLM-based policy verification can provide meaningful support in the loan application review process.

At the same time, the prototype cannot replace a full manual review. Some rules depend on information contained in external systems or require business judgement that is difficult to assess automatically. In addition, the verification agent occasionally produced false-positive findings and tended to adopt a conservative interpretation of policy requirements. While this behaviour increases manual follow-up effort, it is generally preferable to missing potential policy violations.

Overall, the policy checker should be viewed as a decision-support tool rather than a replacement for human expertise. Its main value lies in assisting reviewers by automatically screening loan applications, highlighting potential issues, and drawing attention to policy exceptions and missing information that might otherwise be overlooked.

Several directions for future work remain. The current implementation focuses on a single policy framework and a limited number of loan applications. Future versions should incorporate additional policy documents and related frameworks, allowing the system to determine automatically which policies and institutional levels are relevant in a particular context. Further improvements could include more automated document preprocessing, larger-scale evaluations on additional loan applications, and mechanisms for adapting efficiently to policy updates without manual changes to the system. Finally, cross-rule reasoning and additional summarization layers may help improve consistency, reduce false positives, and provide users with more concise and actionable review results.

\section*{Acknowledgements}

I would like to thank Erste Bank for the opportunity to complete this internship. Special thanks go to my supervisor, Bernhard Pfann, for his support throughout the entire project, to Christina Aschner-Langaschek, who answered all my questions about the lending guidelines and to Andreas Esders-Kopecky for his technical support. Further, I want to thank my PhD supervisor Stefan Szeider his helpful advice throughout  the internship.

\printbibliography

\end{document}